\documentclass[11pt,a4paper]{article}
\usepackage[hyperref]{acl2019}
\usepackage{times}
\usepackage{latexsym}
\usepackage{graphicx}
\usepackage{booktabs}
\usepackage{array}
\usepackage{caption}
\usepackage{tabularx}
\graphicspath{ {./images/} }

\usepackage{url}

\usepackage[utf8]{inputenc}
\usepackage{tikz}
\usetikzlibrary{shapes, arrows.meta}

\aclfinalcopy 

\title{Improving QA Model Performance with Cartographic Inoculation}

\author{Allen Chen \\
  (UT Austin) \\
  \texttt{achen2c@utexas.edu} \\\And
  Okan Tanrikulu \\
  (UT Austin) \\
  \texttt{okan.tanrikulu@utexas.edu} \\}

\date{}

\begin{document}
\maketitle
\begin{abstract}
 QA models are faced with complex and open-ended contextual reasoning problems, but can often learn well-performing solution heuristics by exploiting dataset-specific patterns in their training data. These patterns, or "dataset artifacts", reduce the model's ability to generalize to real-world QA problems. Utilizing an ElectraSmallDiscriminator model trained for QA, we analyze the impacts and incidence of dataset artifacts using an adversarial challenge set designed to confuse models reliant on artifacts for prediction. Extending existing work on methods for mitigating artifact impacts, we propose cartographic inoculation, a novel method that fine-tunes models on an optimized subset of the challenge data to reduce model reliance on dataset artifacts. We show that by selectively fine-tuning a model on ambiguous adversarial examples from a challenge set, significant performance improvements can be made on the full challenge dataset with minimal loss of model generalizability to other challenging environments and QA datasets.
\end{abstract}

\section{Introduction}

Among modern applications of Natural Language Processing (NLP), one of the most widespread and still-growing is Question Answering, or QA, which allows for queries generated with natural language to be parsed and subsequently answered by a language model based on the contextual information available to it \citep{7755228}. Due to the vast number and varied difficulty of possible questions that can be posed, the ability to learn context comprehension techniques that effectively generalize when faced with unfamiliar real-world inputs is arguably the most important function of a Question Answering (QA) language model \citep{CALIJORNESOARES2020635}. 

\begin{figure}[htp]
    \centering
    \includegraphics[width=7cm]{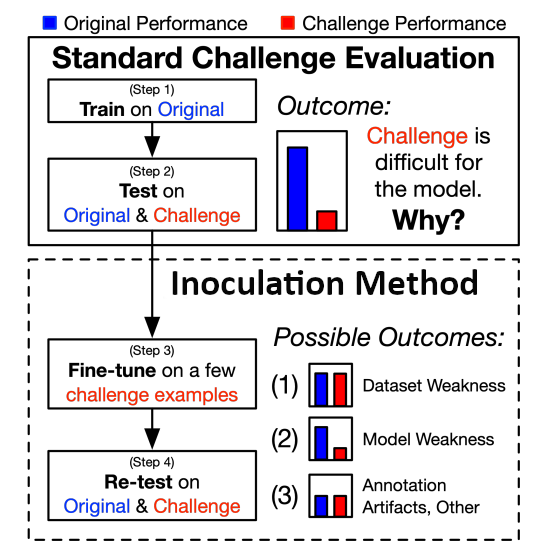}
    \caption{Visualization depicting the inoculation by fine-tuning method and potential outcomes, figure adapted from \citet{liu2019inoculation}}
    \label{fig:inoculationgraphic}
\end{figure}

A significant threat to model generalization are dataset artifacts, human-built patterns in training datasets unlikely to exist in real-world examples. When correlated with answers, these patterns can easily trick QA models into learning maladaptive prediction heuristics \citep{gardner-etal-2021-competency, poliak-etal-2018-hypothesis}. For example, if query construction involves simple permutation of context sentences into questions, as in the QA pair: “The Denver Broncos won the Super Bowl in 1950” $\vert$ ”What year did the Denver Broncos win the Super Bowl in?”, models might maladaptively learn to prioritize information in sentences featuring high lexical overlap with the query, leading to artificially inflated performance on standard benchmarks yet poor real-world applicability.

A promising approach for assessing and reducing model reliance on dataset artifacts is the “Inoculation by Fine-Tuning” method proposed by \citet{liu2019inoculation}, summarized in Figure \ref{fig:inoculationgraphic}. This process utilizes “challenge datasets” containing elements specifically designed to target and reveal model dependencies on non-generalizable heuristics. \citet{liu2019inoculation}'s findings indicate that fine-tuning on a subset of such adversarial examples can enhance a model's resilience against misleading elements, thereby improving its overall performance.

Nevertheless, while the details and subsequent analysis of the performance gains from inoculation have been well-documented in the natural language inference (NLI) setting \citep{liu2019inoculation, gupta2023multiset}, the results for QA are based solely on high-level evaluation metrics and leave much to be desired in the followup analysis. Moreover, the original \citet{liu2019inoculation} study reports a decline in original training set performance post-inoculation, suggesting potential overfitting to the adversarial challenge set in the QA case. This leaves room in the literature for improvements to the standard inoculation methodology and raises critical questions about the effectiveness of QA inoculation techniques: Do they genuinely diminish artifact reliance, or do they inadvertently instill alternative, albeit less detrimental, artifacts?

This paper seeks to address these concerns by conducting an in-depth evaluation of inoculation by fine-tuning's effectiveness at reducing artifact reliance in QA models and proposing a novel variant of the inoculation method that improves upon its weaknesses. We examine 1787 QA instances from model evaluations on the Adversarial SQuAD dataset, breaking down performance gains from inoculation by question-type and query-word. Furthermore, we introduce Random Adversarial SQuAD, a novel challenge set designed to test for Adversarial SQuAD artifacts post-inoculation. Extending the work of \citet{liu2019inoculation}, we propose cartographic inoculation, a method that employs data mapping techniques to strategically fine-tune models on the most ambiguous examples in a challenge set. Our results demonstrate that this hybrid approach significantly surpasses conventional inoculation, yielding more rapid performance gains, enhanced out-of-domain accuracy, and diminished overfitting effects.

\section{Background and Related Work}
\subsection{The QA Task}
In the context of NLP, the general format of a Question Answering (QA) task involves two main inputs: a Question, which is a string containing a query, and a Context passage containing information relevant to answering the query. Given these two inputs, a QA model generates what it believes to be the answer to the given query, often in direct reference to the context passage \citep{7755228}. The ability of a model to accurately pinpoint the answer span depends on its comprehension of both the question's intent and the context's content, making QA tasks a useful benchmark for evaluating the depth of a model's natural language understanding \citep{CALIJORNESOARES2020635}.
\subsection{Dataset Artifacts}
The incidence of dataset artifacts, inherent biases or patterns unintentionally embedded in training data, pose significant challenges in the NLP domain. Numerous studies have attempted to characterize and evaluate dataset artifacts across a variety of NLP tasks and datasets, such as hypothesis-only baselines in NLI \citep{poliak-etal-2018-hypothesis} and question/passage only models for QA \citep{kaushik-lipton-2018-much}. Other studies have developed techniques that minimize the influence of such artifacts, such as contrastive sets \citep{DBLP:journals/corr/abs-2104-08735} and ensemble-based debiasing \citep{he-etal-2019-unlearn}. However, many of these techniques are highly technical and labor-intensive in their implementations. In contrast, our proposed cartographic inoculation method is intuitive and easy-to-use, integrating easily into existing model training and testing frameworks.
\subsection{Adversarial Challenge Sets}
Owing to their relative ease-of-use and effectiveness, adversarial challenge sets have emerged as a common tool for testing a model's reliance on dataset artifacts. These sets introduce adversarial examples designed to confuse models into making incorrect inferences on these examples by exploiting specific decision-making heuristics, like artifact-based cues, utilized by the model \citep{morris-etal-2020-textattack}. While traditionally used as an evaluation technique, we leverage challenge sets as a treatment tool, contributing to the existing literature our findings concerning their effectiveness in this alternative role.  
\subsection{Inoculation by Fine-Tuning}
The "Inoculation by Fine-Tuning" approach proposed by \citet{liu2019inoculation} involves exposing models to a small subset of adversarial examples during training, 'inoculating' them against the specific types of errors or biases exploited by that adversary. While their results convincingly suggest the effectiveness of the inoculation method at improving model performance on challenge sets, follow-up analysis focuses mostly on the NLI classification task and provides only limited insight into the impact of inoculation in the QA context. Furthermore, results for the QA task showed a significant performance dip on the training set post-inoculation, leading \citet{liu2019inoculation} to hypothesize that their fine-tuning may have overfit the resulting model to artifacts in the challenge set. In an attempt to fill this gap in the literature, we dig deeper into the results of inoculation and propose improvements to the inoculation method for the QA task. 
\subsection{Dataset Cartography}
Dataset cartography focuses on mapping the landscape of training datasets to understand the learning dynamics and difficulty associated with individual examples. Pioneered by \citet{swayamdipta2020dataset}, the cartographic approach visualizes datasets based on model confidence and prediction variability during training. This visualization helps to identify easy-to-learn examples that models quickly master, as well as challenging examples that remain ambiguous or hard to learn even after several training epochs. Our proposed "cartographic inoculation" technique leverages insights from dataset cartography to enhance the standard inoculation by fine-tuning method, extending the existing literature from both fields.

\section{Datasets}
We utilize four datasets to facilitate and validate our inoculation methodology, summarized below. For more information, please refer to the original publications.
\subsection{SQuAD}
The Stanford Question Answering Dataset (SQuAD), developed by \citet{rajpurkar2016squad}, is a widely recognized benchmark for machine comprehension, featuring a corpus of over 100,000 questions formulated by crowdworkers from Wikipedia articles. Each entry in SQuAD consists of a context passage, a corresponding question, and an answer span located within the passage. We employ SQuAD to establish a baseline for our model's proficiency in QA tasks.
\subsection{Adversarial SQuAD}
\citet{jia2017adversarial} expanded upon SQuAD with Adversarial SQuAD, which appends adversarial distractor sentences to the ends of context passages across a subset of SQuAD examples \footnote{All references to Adversarial SQuAD in this paper refer specifically to the AddOneSent split of Adversarial SQuAD, which appends only a single distractor to each context passage}. These sentences are crafted to challenge model robustness by including plausible yet incorrect answer choices or subtly misleading information. In this study, we use Adversarial SQuAD both to assess baseline model vulnerabilities and to enhance model performance through inoculation techniques.
\subsection{Randomized Adversarial SQuAD}
To prevent the model from merely adapting to adversarial patterns, we modified the Adversarial SQuAD dataset by shuffling the sentence positions within a context, otherwise leaving everything else intact. We call this novel challenge set 'Randomized Adversarial SQuAD', and utilize it to assess whether the inoculated model overfits to Adversarial SQuAD by simply learning to avoid distractors via the end-of-context positional artifact associated with distracting elements. 
\subsection{TriviaQA}
We extend our evaluation further to TriviaQA, a dataset comprising less structured, open-ended trivia questions from varied sources \citep{joshi2017triviaqa}. TriviaQA serves as a secondary benchmark for testing our inoculated models, offering insights into whether improvements in model performance reflect broader enhancements in model robustness and comprehension skills rather than SQuAD-specific tuning.

\section{Baseline Model Performance on Adversarial SQuAD} 

\begin{figure}[htp]
    \centering
    \includegraphics[width=7cm, height=6cm]{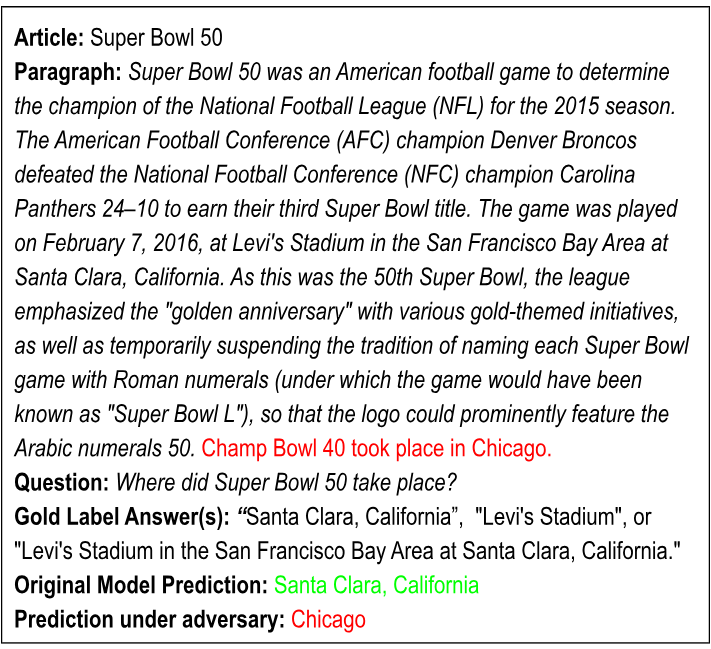}
    \caption{Example of a model error on Adversarial SQuAD, figure adapted from \citet{jia2017adversarial}}
    \label{fig:adversaryexample}
\end{figure}

Utilizing an ElectraSmallDiscriminator model trained for 3 epochs on SQuAD, we achieve a 78\% Exact Match (EM) score, representing the percentage of correctly classified examples, and an 86\% F1 score on the SQuAD dataset. Performance on Adversarial SQuAD is much lower, at 62.70\% EM and 71.02\% F1. In order to understand this decline, we examine 1,787 examples from Adversarial SQuAD, comparing model predictions to ground truth answers and aggregating examples by question type and query-word. 

Adversarial SQuAD undermines model performance by appending distractor sentences designed to mislead model predictions to the end of context passages. These distracting examples are interspersed throughout the challenge set, and are expected to challenge our model the most. We identify 728 distracting examples in Adversarial SQuAD and find our model fails to answer more than half of them, outputting a mere 45.3\% EM accuracy on distracting examples as opposed to 74.7\% on non-distracting examples. See Table \ref{tab:performance_comparison} for more information. 

Observing individual errors, we examine why these distracting examples effectively confuse our model. Consider Figure \ref{fig:adversaryexample}, which shows model output on an Adversarial SQuAD example before and after the addition of a distractor sentence. In the SQuAD version of the example, the model correctly extracts the answer from the context passage, but when including an adversarial element, the prediction now incorrectly focuses on the information contained in the distractor. Note how the distractor sentence from Figure \ref{fig:adversaryexample}, “Champ Bowl 40 took place in Chicago” has a high lexical overlap with the question, “Where did the Super Bowl 50 take place?”.

\begin{table}[htbp]
\centering
\caption{Model performance on distracting versus non-distracting examples}
\label{tab:performance_comparison}
\begin{tabular}{@{}lrrr@{}}
\toprule
\textbf{} & \textbf{Distracting} & \textbf{Non-Distracting} & \textbf{Total} \\ 
\midrule
Total     & 728                  & 1059             & 1787           \\
Correct   & 329                  & \textbf{791}              & 1120           \\
Incorrect     & \textbf{399}                  & 268              & 667            \\
\bottomrule
\end{tabular}
\end{table}

Further examination shows that misclassified examples consistently have high syntactic similarity between questions and distractors, suggesting that our model may rely on \textit{lexical overlap artifacts} from SQuAD as a heuristic for locating answers.

Intuitively, this makes sense – in order to create the examples associated with SQuAD, human crowdworkers were asked to read Wikipedia-sourced context passages en masse and generate questions whose answers were contained within those passages \citep{rajpurkar2016squad}. A simple way to do this would be to find a fact in the passage, such as “The color of the sky is blue”, and rephrase it as a question, “What is the color of the sky?”. Thus, a model could maladaptively learn that the answers to its questions are most often contained in parts of the passage worded similarly to the question, bypassing the need for a more complex comprehension of the information contained in the context. In order to improve our baseline model’s performance on distracting examples, we aim to reduce its reliance on lexical overlap artifacts during inference.

\begin{table}[htbp]
\centering
\caption{Distribution of question types and query-words in Adversarial SQuAD.}
\label{tab:question_distribution}
\begin{tabular}{@{}lllr@{}}
\toprule
\textbf{Question Type} & \textbf{Query Word} & \textbf{POS Tag} & \textbf{Count} \\ 
\midrule
Descriptive & how    & WRB &  76 \\
Descriptive & what   & WDT &   6 \\
Descriptive & what   & WP  & 1001 \\
Descriptive & which  & JJ  &  41 \\
Descriptive & which  & NNP &  22 \\
Descriptive & which  & WDT &  63 \\
Explanatory & when   & WRB & 143 \\
Explanatory & where  & WRB &  74 \\
Explanatory & why    & WRB &  34 \\
Other       & (NA)   & (NA)&  20 \\
Person      & who    & NNP &   3 \\
Person      & who    & WP  & 182 \\
Quantitative& how    & WRB & 122 \\
\bottomrule
\end{tabular}
\end{table}

Nevertheless, prediction accuracy on distracting examples alone is not wholly reflective of our model’s overall performance. To support deeper analysis, we annotate category labels and computed statistics onto all 1787 of our Adversarial SQuAD examples. We identify four main categories of question: Descriptive, Explanatory, Person, and Quantitative. Within each question-type category, we list the query-words associated with its examples, resulting in the taxonomy shown in Table \ref{tab:question_distribution}. Descriptive "what" questions strongly outweigh all other categories, comprising 56\% of the dataset, but other than that, we find no significant outliers. 

With our dataset thus classified, we begin to examine the range and type of errors made by the baseline model on the Adversarial SQuAD dataset. Table \ref{tab:baseline_accuracy} summarizes baseline model accuracy by question type, revealing that the model performs worst on “why” (explanatory) and “how” (descriptive/quantitative) questions. This is within expectation, as questions of this nature ought to require a more comprehensive understanding of the context which can identify relationships between causes and effects, and are not likely to exhibit well-defined patterns in their structure (such as the implicit association between the query-word “where” and an answer referencing a location). Evaluating our improved model’s performance on these harder categories of query-words will allow us to ensure performance gains reflect an enhancement of the model’s overall understanding of the QA paradigm and not improvement on a dominant but potentially simple-to-learn category of question, such as "what". 

\begin{table}[htbp]
\centering
\caption{Baseline Accuracy (EM) across different query words}
\label{tab:baseline_accuracy}
\begin{tabular}{@{}lcr@{}}
\toprule
\textbf{Query Word} & \textbf{Count} & \textbf{Accuracy} \\ 
\midrule
ALL    & 1,787 & 62.7\% \\
why    & 34    & 32.4\% \\
how    & 202   & 57.9\% \\
where  & 72    & 62.5\% \\
which  & 136   & 62.5\% \\
what   & 1,026 & 63.7\% \\
who    & 193   & 65.3\% \\
when   & 104   & 68.3\% \\
\bottomrule
\end{tabular}
\end{table}

\section{Improvement Approach}

\begin{figure*}[t]
    \centering
    \includegraphics[width=\textwidth]{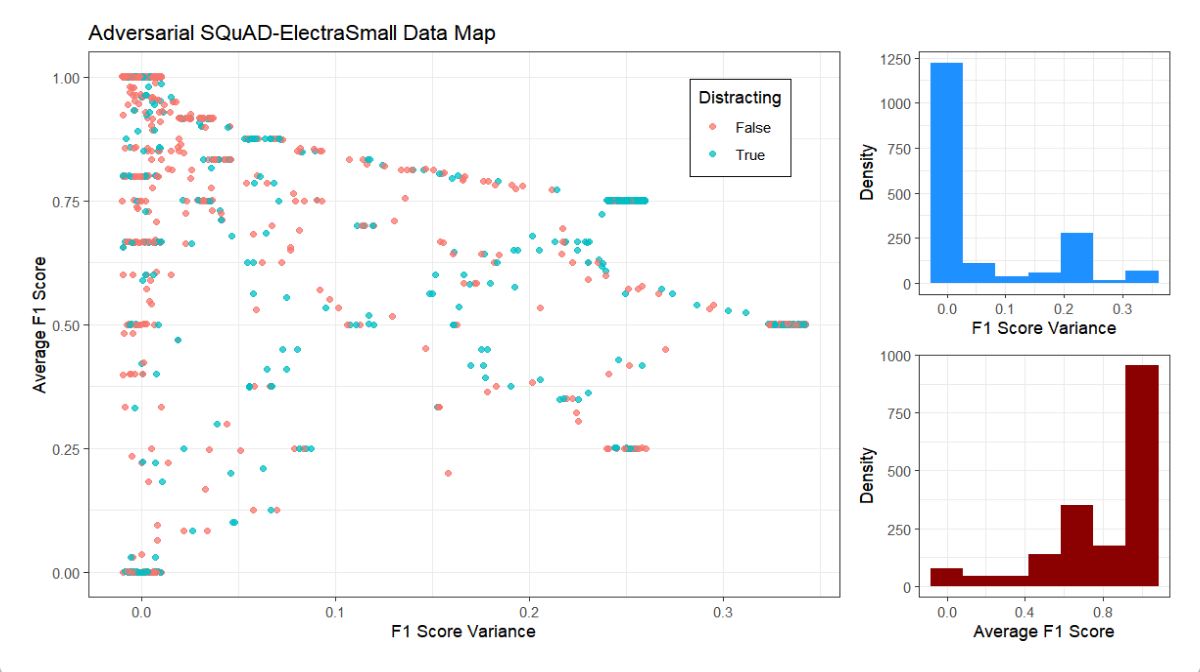}
    \caption{Data map for the Adversarial SQuAD challenge set. Note the high F1 variance (indicating ambiguity) on many distracting examples.}
    \label{fig:datamap}
\end{figure*}

In analyzing our baseline model's performance, we identified two improvement goals: enhancing performance on Adversarial SQuAD's distracting examples by reducing model reliance on the lexical overlap artifact, and ensuring accuracy gains span various question-types and query-words, especially in complex "why" and "how" questions. To achieve these goals, we employ the "Inoculation by Fine-Tuning" method by \citet{liu2019inoculation}, using Adversarial SQuAD as training data to heighten resistance to adversarial elements.
\subsection{Inoculation Approach}
The primary driver of model improvement in the inoculation method described is the "inoculation set", a subset of examples from the adversarial challenge set that an uninoculated model is trained on so it can adapt to and overcome adversarial elements. The standard procedure entails initially assessing the model on both the training and an adversarial set, followed by fine-tuning on randomly sampled examples from the adversarial set and then re-evaluating. Ideally, the performance gap on the training and challenge sets should converge to zero as the size of the inoculation set grows, though performance on the training set may be damaged in the case that elements of the challenge set (such as gold-label distribution, presence or lack of specific dataset artifacts, etc.) consistently contradict the original training data \citep{liu2019inoculation}.

We expanded on the standard inoculation method by testing our inoculated model on additional datasets and exploring specialized methods for selecting the inoculation set. We compared three inoculation set selection methods: random sampling from the full challenge dataset, sampling only distracting examples, and selecting challenging examples using dataset cartography. Performance was then assessed on SQuAD, Adversarial SQuAD, TriviaQA, and our hand-made "Random Adversarial SQuAD", ensuring that inoculation results generalize to both in and out-of domain QA benchmarks and adversarial environments. 
\subsection{Loading the Datasets}
We load the SQuAD and Adversarial SQuAD datasets using the HuggingFace ‘transformers’\footnote{https://huggingface.co/docs/transformers/index} package. To build Randomized Adversarial SQuAD, we extract Adversarial SQuAD and randomize the ordering of each example’s context passage sentences. We download TriviaQA from the University of Washington website \footnote{https://nlp.cs.washington.edu/triviaqa/} and modify its structure to work with our model training and testing pipelines, excluding examples without answers.
\subsection{Cartographic Inoculation}
\begin{figure}[htp]
    \centering
    \includegraphics[width=7cm]{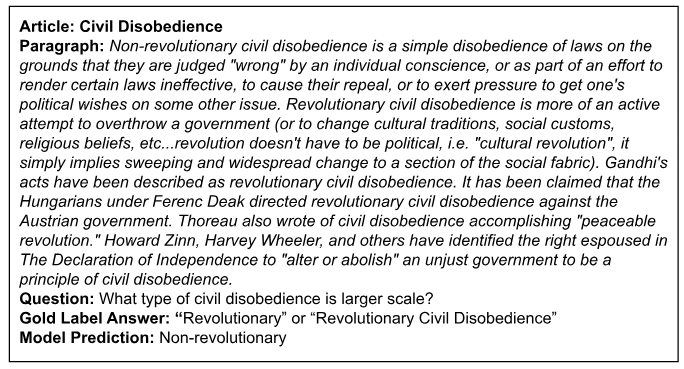}
    \caption{Example of an ambiguous example as identified by dataset cartography. Note the complexity of the context passage and open-ended nature of the question.}
    \label{fig:ambiguousexample}
\end{figure}

Implementing random sampling for the standard and distractors-only inoculation sets was straightforward, but using dataset cartography for inoculation set selection required us to adapt \citet{swayamdipta2020dataset}'s methods for the QA task. Whereas the original paper uses confidence and variance metrics based on classification probabilities to plot examples for the NLI classification task, we use the text-similarity F1 score \footnote{See \citet{rajpurkar2016squad} for more details.} between the gold label answer and the model’s predicted answer as a measure of accuracy for each example. Aggregating results from 5 training epochs of our baseline model on the Adversarial set, we compute the average F1 score and its variance for each example and plot them in Figure \ref{fig:datamap} to identify the most ambiguous (highest-variance)\footnote{For more information about how ambiguous examples are defined, see \citet{swayamdipta2020dataset}} examples. Figure \ref{fig:ambiguousexample} shows one such example. Unlike our other two styles, which choose their subsets randomly from the sample provided, this method is deterministic, using the $N$ most ambiguous examples in each inoculation set. 
\subsection{Evaluation}
Our evaluation of the inoculated models follows \citet{liu2019inoculation}'s framework. First, we evaluate our baseline model on the four datasets. Each inoculation style is then applied by fine-tuning the baseline model for a single epoch on varying sizes of the inoculation set, starting with 100 examples and increasing incrementally up to 700, with the model reinoculated and reevaluated at each stage. TriviaQA evaluations are reserved for baseline and final model iterations.

\begin{figure*}[t]
    \centering
    \includegraphics[width=\textwidth]{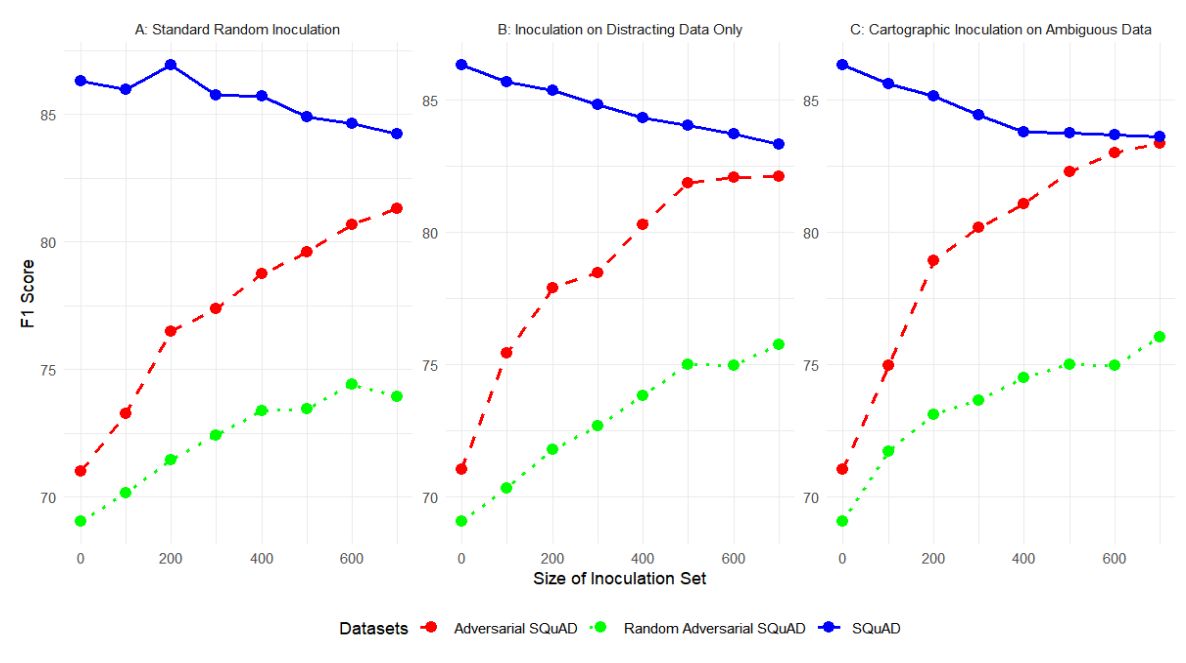}
    \caption{Results of inoculation. Plotting EM scores instead of F1 scores produces similar trends.}
    \label{fig:results}
\end{figure*}

\section{Results}

Figure \ref{fig:results} displays inoculation outcomes using the simply random baseline, distractors-only, and dataset cartography selection strategies across various inoculation set sizes. Each method effectively narrows the performance gap between SQuAD and Adversarial SQuAD, with a slight performance decline on SQuAD, matching the general trend observed by \citet{liu2019inoculation}. 

Among these strategies, dataset cartography-based inoculation excels, reducing the F1 performance gap to \textless1.5\% with only 500 examples and demonstrating superior performance on both challenge and non-challenge datasets. Table \ref{tab:inoculation_performance} contrasts the top-performing models for each method, underscoring the efficacy of cartographic inoculation.

\subsection{Performance on the Random Adversary}
The performance of standard inoculation on Random Adversarial SQuAD plateaus and eventually declines with larger samples of the inoculation set, lending credibility to \citet{liu2019inoculation}’s hypothesis that their inoculation method was overfitting to Adversarial SQuAD by learning to avoid distractors based on the end-of-context positional artifact. However, on the distractor-only and cartographic selection techniques, performance on the random adversary increases steadily. This result, which merits further exploration, suggests that these modifications of the general inoculation technique might counteract overfitting while enhancing model performance.
\subsection{Performance on Adversarial SQuAD}
Using our final 500-example cartographically inoculated model, we analyze the impacts of inoculation on a more granular scale. Breaking down our performance gains, we find that inoculation manages to improve performance on both distracting and non-distracting examples by 24.9\% and 3\%, respectively. And as Table \ref{tab:model_accuracy_change} shows, cartographic inoculation enhances performance across all example categories. Reassuringly, we observe accuracy gains on "why" and “how” questions, which were previously identified as difficult query-word categories in the data. These improvements suggest that our model has gained proficiency in comprehending complex, open-ended questions, possibly as a result of being fine-tuned on ambiguous challenge set examples. 

\begin{table}[htbp]
\centering
\caption{Cartographically-inoculated model (500 examples) accuracy on Adversarial SQuAD, by question type.}
\label{tab:model_accuracy_change}
\begin{tabular}{@{}p{1cm}p{2cm}p{2cm}p{1.5cm}@{}}
\toprule
\textbf{Query Word} & \textbf{Baseline Accuracy} & \textbf{Cartography Accuracy} & \textbf{Change} \\
\midrule
ALL    & 62.7\% & 74.6\% & +11.9\% \\
why    & 32.4\% & 41.2\% & \textbf{+8.8}\% \\
how    & 57.9\% & 74.0\% & \textbf{+16.1\%} \\
where  & 62.5\% & 73.8\% & +11.3\% \\
which  & 62.5\% & 72.1\% & +9.6\% \\
what   & 63.7\% & 74.3\% & +10.6\% \\
who    & 65.3\% & 79.9\% & +14.6\% \\
when   & 68.3\% & 86.5\% & +18.2\% \\
\bottomrule
\end{tabular}
\end{table}

\begin{table*}[htbp]
\centering
\caption{Model Performance under different inoculation methods.}
\label{tab:inoculation_performance}
\begin{tabular}{@{}lcccccccc@{}}
\toprule
 & \multicolumn{2}{c}{\textbf{SQuAD}} & \multicolumn{2}{c}{\textbf{Adv SQuAD}} & \multicolumn{2}{c}{\textbf{Rand Adv SQuAD}} & \multicolumn{2}{c}{\textbf{TriviaQA}} \\
\cmidrule(r){2-3} \cmidrule(lr){4-5} \cmidrule(lr){6-7} \cmidrule(l){8-9}
\textbf{Method} & \textbf{EM} & \textbf{F1} & \textbf{EM} & \textbf{F1} & \textbf{EM} & \textbf{F1} & \textbf{EM} & \textbf{F1} \\
\midrule
No Inoculation (Baseline) & \textbf{78.54} & \textbf{86.32} & 63.63 & 71.02 & 61.55 & 69.05 & 42.35 & 49.84 \\
Standard Inoculation (N=700) & 76.39 & 84.24 & 74.6 & 81.33 & 66.98 & 73.93 & 40.5 & 48 \\
Distractors-Only Inoculation (N=700) & 75.53 & 83.32 & 75.43 & 82.13 & \textbf{68.27} & \textbf{75.73} & 38.17 & 45.35 \\
Cartographic Inoculation (N=500) & 75.78 & 83.611 & \textbf{76.89} & \textbf{83.38} & \textbf{68.16} & \textbf{76.03} & 39.58 & 46.9 \\
\bottomrule
\end{tabular}
\end{table*}

\section{Discussion}
In reducing our standard SQuAD-based QA model's reliance on dataset artifacts, our goal was to improve our model’s performance on both distracting examples (without simply overfitting to the challenge set) and on the hardest question categories of our challenge set. Based on our final inoculated model’s robust performance on both Adversarial SQuAD, it appears that we were successful in achieving these goals. Our novel cartographic inoculation method showed itself to be capable of closing the performance gap between the original training set and its adversarial counterpart faster and at a higher performance level than the standard method, while still managing to effectively generalize to other challenge sets. 

We suspect that by targeting ambiguous examples, the model learns more efficiently by gaining deeper and more actionable insights from difficult parts of the dataset. And since a model fine-tuned in this way does not learn to solely focus on distracting examples, performance on the original training set is less degraded by any overfitting to challenge set features. Furthermore, performance improvements on Random Adversarial SQuAD suggest that the positive effects of inoculation on Adversarial SQuAD are not dependent on the distractors being \textit{positionally-invariant}, as \citet{liu2019inoculation} hypothesized they might be. The result is a total upgrade to the original inoculation technique which improves upon its strengths while simultaneously remediating many of its projected weaknesses.
\subsection{Limitations}
Our cartographic inoculation method assumes the overlap between the presence of adversarial elements and ambiguity is high, as observed in the data map for Adversarial SQuAD. When adversarial examples are not highly ambiguous, inoculation on ambiguous data may end up being self-defeating, as an inoculation set devoid of any adversarial elements could be generated, leading to the model simply fine-tuning on ambiguous examples from the original training data. Nonetheless, we believe the general process of using data cartography techniques to select inoculation sets should remain effective even if the exact heuristics by which selection is carried out change. 

\section{Conclusion}
The incidence of dataset artifacts within benchmark QA datasets like SQuAD disguises a model’s true performance in generalizing to real-world data, degrading the value of such models in practical applications. 

In this work, we presented cartographic inoculation, a new approach to mitigating the impacts of certain SQuAD-based dataset artifacts that builds on the inoculation by fine-tuning method from \citet{liu2019inoculation} by using dataset cartography techniques to identify ambiguous examples in Adversarial SQuAD for inoculation set selection. Addressing concerns from \citet{liu2019inoculation} that the standard inoculation method on Adversarial SQuAD may be overfitting to the challenge set, we validate the effectiveness of our method with both a hand-made adversary designed to test the inoculated model’s dependence on positional artifacts in Adversarial SQuAD and a detailed analysis of model performance before and after inoculation. 

Relative to both the standard inoculation strategy using random sampling and a distractors-only strategy which selected only adversarial examples, our novel method raises model performance on the challenge set to match its performance on the original training data using fewer examples, with less degradation of in and out-of domain benchmark performance, and without loss of generalizability to other challenging environments. 

\bibliography{acl2019}
\bibliographystyle{acl_natbib}

\end{document}